\documentclass[11pt]{scrartcl}
\DeclareOldFontCommand{\rm}{\normalfont\rmfamily}{\mathrm}
\DeclareOldFontCommand{\sf}{\normalfont\sffamily}{\mathsf}
\DeclareOldFontCommand{\tt}{\normalfont\ttfamily}{\mathtt}
\DeclareOldFontCommand{\bf}{\normalfont\bfseries}{\mathbf}
\DeclareOldFontCommand{\it}{\normalfont\itshape}{\mathit}
\DeclareOldFontCommand{\sl}{\normalfont\slshape}{\@nomath\sl}
\DeclareOldFontCommand{\sc}{\normalfont\scshape}{\@nomath\sc}
\DeclareRobustCommand*\cal{\@fontswitch\relax\mathcal}
\DeclareRobustCommand*\mit{\@fontswitch\relax\mathnormal}

\usepackage{titling}
\predate{}
\postdate{}

\usepackage{graphicx}
\usepackage{amsmath,amssymb,bm,amsthm,physics}

\usepackage[numbers,sort&compress]{natbib}

\usepackage[usenames,svgnames]{xcolor}
\usepackage[setpagesize=false,colorlinks=true,linkcolor=blue,citecolor=blue,breaklinks=true]{hyperref}
\usepackage{authblk}
\usepackage{cleveref}

\begin{document}
\nocite{apsrev41control}

\title{\Large Interplay between depth of neural networks and locality of target functions}
\author[1]{\normalsize Takashi Mori}
\author[1,2,3]{\normalsize Masahito Ueda}
\affil[1]{\it RIKEN Center for Emergent Matter Science (CEMS), Wako 351-0198, Japan}
\affil[2]{\it Department of Physics, The University of Tokyo, Bunkyo-ku, Tokyo 113-0033, Japan}
\affil[3]{\it Institute for Physics of Intelligence, University of Tokyo, Bunkyo-ku, Tokyo 113-0033, Japan}
\date{}

\maketitle
\begin{abstract}
It has been recognized that heavily overparameterized deep neural networks (DNNs) exhibit surprisingly good generalization performance in various machine-learning tasks.
Although benefits of depth have been investigated from different perspectives such as the approximation theory and the statistical learning theory, existing theories do not adequately explain the empirical success of overparameterized DNNs.
In this work, we report a remarkable interplay between depth and locality of a target function.
We introduce $k$-local and $k$-global functions, and find that depth is beneficial for learning local functions but detrimental to learning global functions.
This interplay is not properly captured by the neural tangent kernel, which describes an infinitely wide neural network within the lazy learning regime. 
\end{abstract}

\section{Introduction}
\label{sec:intro}
Deep neural networks (DNNs) have achieved unparalleled success in various tasks of artificial intelligence such as image classification~\citep{Krizhevsky2012,LeCun2015} and speech recognition~\citep{Hinton2012}.
Empirically, DNNs often outperform other machine learning methods such as kernel methods and Gaussian processes, but little is known about the underlying mechanism of outstanding performance of DNNs.

To elucidate benefits of depth, numerous studies have investigated properties of DNNs from various perspectives.
The approximation theory focuses on the expressive power of DNNs~\citep{Yarotsky2017}.
Although the universal approximation theorem states that a sufficiently wide neural network with a single hidden layer can approximate any continuous functions, expressivity of a DNN grows exponentially with increasing the depth rather than the width~\citep{Telgarsky2016, Poole2016, Bianchini2014, Montufar2014}.
In statistical learning theory, the decay rate of the generalization error in large sample asymptotics has been analyzed.
For learning generic smooth functions, shallow networks or other standard methods with linear estimators such as kernel methods already give the optimal rate~\citep{Tsybakov_text}, and hence benefits of depth are not obvious.
On the other hand, for learning smooth functions with some special properties such as the hierarchical compositional property~\citep{Schmidt-HIeber2020} and spatial inhomogeneity of smoothness~\citep{Suzuki2019}, or for learning a certain class of non-smooth functions~\citep{Imaizumi2019}, it has been shown that DNNs show faster decay rates of the generalization error compared with linear estimators.

Although those existing theoretical efforts have revealed interesting and nontrivial properties of DNNs, they do not adequately explain the empirical success of deep learning.
Crucially, in modern machine learning applications, impressive generalization performance has been observed in an overparameterized regime, in which the number of parameters in the network greatly exceeds the number of training data samples~\citep{Zhang2017, Spigler2019, Belkin2019}.
The asymptotic decay rate of the generalization error, which has been studied in statistical learning theory, does not cover an overparameterized regime.
As for the approximation theory, it is far from clear whether high expressive power of DNNs are really beneficial in practical applications~\citep{Ba2014, Eldan2016, Safran2017, Safran2019, Becker2020}.
A recent work~\citep{Malach2019} has demonstrated that a DNN trained by a gradient-based optimization algorithm can only learn functions that are well approximated by a shallow network, indicating that benefits of depth are not due to high expressivity of DNNs.
Thus, benefits of depth for generalization ability of overparameterized DNNs still remain elusive.

In this work, we numerically investigate the effect of depth in learning simple functions, for which no evidence for benefits of depth is found in existing theories.
We here focus on the locality property of target functions, and introduce $k$-local and $k$-global functions.
A $k$-local function is given by a product of pre-fixed $k$ entries of the input vector, whereas a $k$-global function is defined as a global sum of $k$-local functions (we will later consider more general target functions in \cref{sec:general}).
We find that \emph{depth is beneficial for learning $k$-local functions but rather detrimental to learning $k$-global functions}.

We also show that the effect of depth is not correctly captured by theory of the neural tangent kernel (NTK)~\citep{Jacot2018}, which describes an infinitely wide neural network optimized by stochastic gradient descent (SGD) with an infinitesimal learning rate.
Since the NTK is involved with lazy learning regime~\citep{Chizat2019}, in which network parameters stay close to their initial values, the failure of the NTK in capturing the effect of depth implies the importance of feature learning, in which parameters change to learn relevant features.

\subsection{Our contribution}
We summarize our contribution below.
\begin{itemize}
\item We find that benefits of depth in an overparameterized regime are present even for very simple functions such as $k$-local ones (\cref{sec:depth}).
Although it is sometimes emphasized that DNNs can express complex functions, this result shows that benefits of depth are not solely attributed to high expressivity.
\item We find that depth is not always beneficial as is clearly demonstrated for learning $k$-global functions (\cref{sec:depth}).
\item Those results are not explained by the NTK, which describes the lazy learning regime (\cref{sec:depth}).
This fact implies the importance of the feature learning regime, which corresponds to large learning rates (\cref{sec:lr}).
\item The opposite depth dependence of $k$-local and $k$-global functions is also observed for noisy labels (\cref{sec:noise}), the classification task with the cross-entropy loss (\cref{sec:class}), and more general local and global functions (\cref{sec:general}).
Thus, our results are robust.
\end{itemize}

\section{Setup}

We consider supervised learning of a target function $f: \mathbb{R}^d\to\mathbb{R}$ with a training dataset $\mathcal{D}=\qty{\qty(x^{(\mu)},y^{(\mu)}):\mu=1,2,\dots,N}$, where $x^{(\mu)}\in\mathbb{R}^d$ is a $d$-dimensional input data and $y^{(\mu)}=f(x^{(\mu)})$ is its label.
Each input vector $x$ is assumed to be a $d$-dimensional random Gaussian vector: $x\sim\mathcal{N}(0,I_d)$, where $\mathcal{N}(m,\sigma^2)$ denotes the Gaussian distribution of mean $m$ and covariance $\sigma^2$, and $I_d$ denotes the $d$-dimensional identity matrix.
We mainly consider noiseless data, but we will consider noisy labels in \cref{sec:noise}.

In the following, we summarize the setup of our experiments.

\subsection{Target functions}
In this work, instead of developing a general mathematical theory for a wide class of target functions $f$, we show experimental results for concrete target functions.
We focus on the locality of target functions, and introduce $k$-local and $k$-global functions.
Let us randomly fix $k$ integers $\{i_1,i_2,\dots,i_k\}$ with $1\leq i_1<i_2<\dots<i_k\leq d$.\footnote{For fully connected neural networks (FNNs) considered in this paper, without loss of generality, we can choose $i_1=1$, $i_2=2$,\dots, $i_k=k$ because of the permutation symmetry of indices of input vectors.}
A $k$-local function is then defined as
\begin{align}
\text{($k$-local)}\quad f(x)=x_{i_1}x_{i_2}\dots x_{i_k},
\label{eq:k-local}
\end{align}
i.e., a product of the $k$ entries of $x$.
This function is ``local'' in the sense that it depends only on the $k$ entries of the input data\footnote{This property may also be called ``sparsity'' rather than ``locality''. However, in this work, we say that such a function is local as opposed to ``global'' functions.} (we consider the case of $k\ll d$).
On the other hand, a $k$-global function is defined by a global sum of $k$-local functions as follows:
\begin{align}
\text{($k$-global)}\quad f(x)=\frac{1}{\sqrt{d}}\sum_{j=1}^dx_{j+i_1}x_{j+i_2}\dots x_{j+i_k},
\label{eq:k-global}
\end{align}
where we impose the periodic boundary condition $x_{d+i}=x_i$.
The scaling of $1/\sqrt{d}$ is introduced to make typical values of $f(x)$ for $x\sim\mathcal{N}(0,I_d)$ independent of $d$.
In contrast to $k$-local functions, every component of $x$ equally contributes to $k$-global functions.

\subsection{Network architecture}
In this work, we consider fully connected neural networks (FNNs) with $L$ hidden layers, each of which has $h$ nodes.
We call $L$ and $h$ depth and width of the network, respectively.
Weights and biases of the $\ell$th layer are respectively denoted by $w^{(\ell)}$ and $b^{(\ell)}$, and let us denote by $w$ the set of all the weights and biases in the network.
The output of the network $\hat{f}(x,w)\in\mathbb{R}$ is determined as follows:
$\hat{f}(x,w)=w^{(L+1)}z^{(L)}$, $z^{(\ell)}=\varphi\qty(w^{(\ell)}z^{(\ell-1)}+b^{(\ell-1)})$ for $\ell=1,2,\dots,L$, and $z^{(0)}=x$, where $z^{(\ell)}$ is the output of the $\ell$th layer and $\varphi(x)=\max\{x,0\}$ is the component-wise ReLU activation function.

We fix the number of parameters for different depths.
In comparing the performance for different $L$, we fix the number $P$ of parameters.
Since the number of parameters is roughly given by $dh+(L-1)h^2$, $h$ is determined for a given $L$ as the closest integer satisfying 
\begin{align}
P=dh+(L-1)h^2.
\end{align}
In this work, we focus on an overparameterized regime $N\ll P$, in which DNNs empirically show astonishing generalization performance~\citep{Zhang2017, Spigler2019, Belkin2019}.

\subsection{Training procedure}
The network parameters $w$ are adjusted to fit training data samples through minimization of the loss function
\begin{align}
L(w)=\frac{1}{N}\sum_{\mu=1}^N\qty(\hat{f}(x^{(\mu)},w)-f(x^{(\mu)}))^2,
\end{align}
which is nothing but the mean-squared error.
The training of the network is carried out by the SGD
\begin{align}
w_{t+1}=w_t-\eta\nabla_wL_{\mathcal{B}_t}(w_t)
\end{align}
with the learning rate $\eta$ and the mini-batch size $B$ (we fix $B=50$ throughout the paper), where $\mathcal{B}_t\subset\{1,2,\dots,N\}$ satisfying $|\mathcal{B}_t|=B$ denotes the mini-batch at $t$th step and 
\begin{align}
L_{\mathcal{B}_t}(w_t)=\frac{1}{B}\sum_{\mu\in\mathcal{B}_t}\qty(\hat{f}(x^{(\mu)},w_t)-f(x^{(\mu)}))^2
\end{align}
denotes the mini-batch loss.

Biases are initialized at zero, and weights are initialized using the Glorot initialization~\citep{Glorot2010}.
For every 50 epochs, we measure the loss function and stop the training if the measured value falls below $10^{-4}$.
We checked that our conclusion is not sensitive to the threshold value for stopping the training.

Before the training, we first perform the 10-fold cross validation to optimize the learning rate under the Bayesian optimization method~\citep{Snoek2012} (we used the package provided by \citet{Nogueira_github}).
We then train the network via the SGD with the optimized $\eta$.
The generalization performance of the trained network is measured by computing the test error
\begin{align}
\varepsilon_g=\frac{1}{N_\mathrm{test}}\sum_{\mu=1}^{N_\mathrm{test}}\qty(\hat{f}(x'^{(\mu)},w^*)-f(x'^{(\mu)}))^2,
\label{eq:error}
\end{align} 
where $w^*$ denotes the parameters of the trained network, and $\mathcal{D}_\mathrm{test}=\{(x'^{(\mu)},y'^{(\mu)}):\mu=1,2,\dots,N_\mathrm{test}\}$ is a test dataset independent of the training dataset $\mathcal{D}$, where $x'^{(\mu)}\sim\mathcal{N}(0,I_d)$ and $y'^{(\mu)}=f(x'^{(\mu)})$.
Throughout the paper, we set $N_\mathrm{test}=10^5$.

\subsection{Neural tangent kernel}
Following \citet{Arora2019} and \citet{Cao2019}, let us consider a FNN of depth $L$ and width $h$ whose biases $\qty{b^{(\ell)}}$ and weights $\qty{w^{(\ell)}}$ are randomly initialized as $b_i^{(\ell)}=\beta\tilde{b}_i^{(\ell)}$ with $\tilde{b}_i^{(\ell)}\sim\mathcal{N}(0,1)$ and $w_{ij}^{(\ell)}=\sqrt{2/n_{\ell-1}}\tilde{w}_{ij}^{(\ell)}$ with $\tilde{w}_{ij}^{(\ell)}\sim\mathcal{N}(0,1)$ for every $\ell$, where $n_\ell$ is the number of nodes in the $\ell$th layer, i.e. $n_0=d$, $n_1=n_2=\dots=n_L=h$.
The parameter $\beta$ controls the impact of bias terms, and we follow \citet{Jacot2018} to set $\beta=0.1$ in our numerical experiments.
Let us denote by $\tilde{w}$ the set of all the scaled weights $\{\tilde{w}^{(\ell)}\}$ and biases $\{\tilde{b}^{(\ell)}\}$.
The network output is written as $f(x,\tilde{w})$.

When the network is sufficiently wide and the learning rate $\eta$ for $w$ is sufficiently small\footnote{Here we remark that the scaled learning rate $\tilde{\eta}$ for $\tilde{w}$ can be finite in the large-width limit~\citep{Lee2019, Lewkowycz2020}.
This means that the original learning rate $\eta$ for $w$ should be proportional to $1/h$ in order to enter the NTK regime.}, the scaled parameters $\tilde{w}$ stay close to their random initialized values $\tilde{w}_0$ during training, and hence $f(x,\tilde{w})$ is approximated by a linear function of $\tilde{w}-\tilde{w}_0$:
\begin{align}
f(x,\tilde{w})=f(x,\tilde{w}_0)+\left.\nabla_{\tilde{w}}f(x,\tilde{w})\right|_{\tilde{w}=\tilde{w}_0}\cdot(\tilde{w}-\tilde{w}_0).
\end{align}
As a result, the minimization of the loss function is equivalent to the kernel regression with the NTK defined as
\begin{align}
\Theta^{(L)}(x,x')=\lim_{h\to\infty}\mathbb{E}_{\tilde{w}}\qty[\nabla_{\tilde{w}}f(x,\tilde{w})^\top\nabla_{\tilde{w}}f(x,\tilde{w})],
\end{align}
where $\mathbb{E}_{\tilde{w}}$ denotes the average over random initializations of $\tilde{w}$.
By using the ReLU activation, we can give an explicit expression of the NTK that is suited for numerical calculations.
See Appendix~\ref{sec:formula_NTK} for the detail.

It is shown that the minimization of the loss function using the NTK yields the output function
\begin{align}
f^{\mathrm{NTK}}(x)=\sum_{\mu,\nu=1}^N\Theta^{(L)}(x,x^{(\mu)})\qty(K^{-1})_{\mu,\nu}y^{(\nu)},
\end{align}
where $K^{-1}$ is the inverse matrix of the Gram matrix $K_{\mu,\nu}=\Theta^{(L)}(x^{(\mu)},x^{(\nu)})$.

\begin{figure}[t]
\centering
\includegraphics[width=0.45\linewidth]{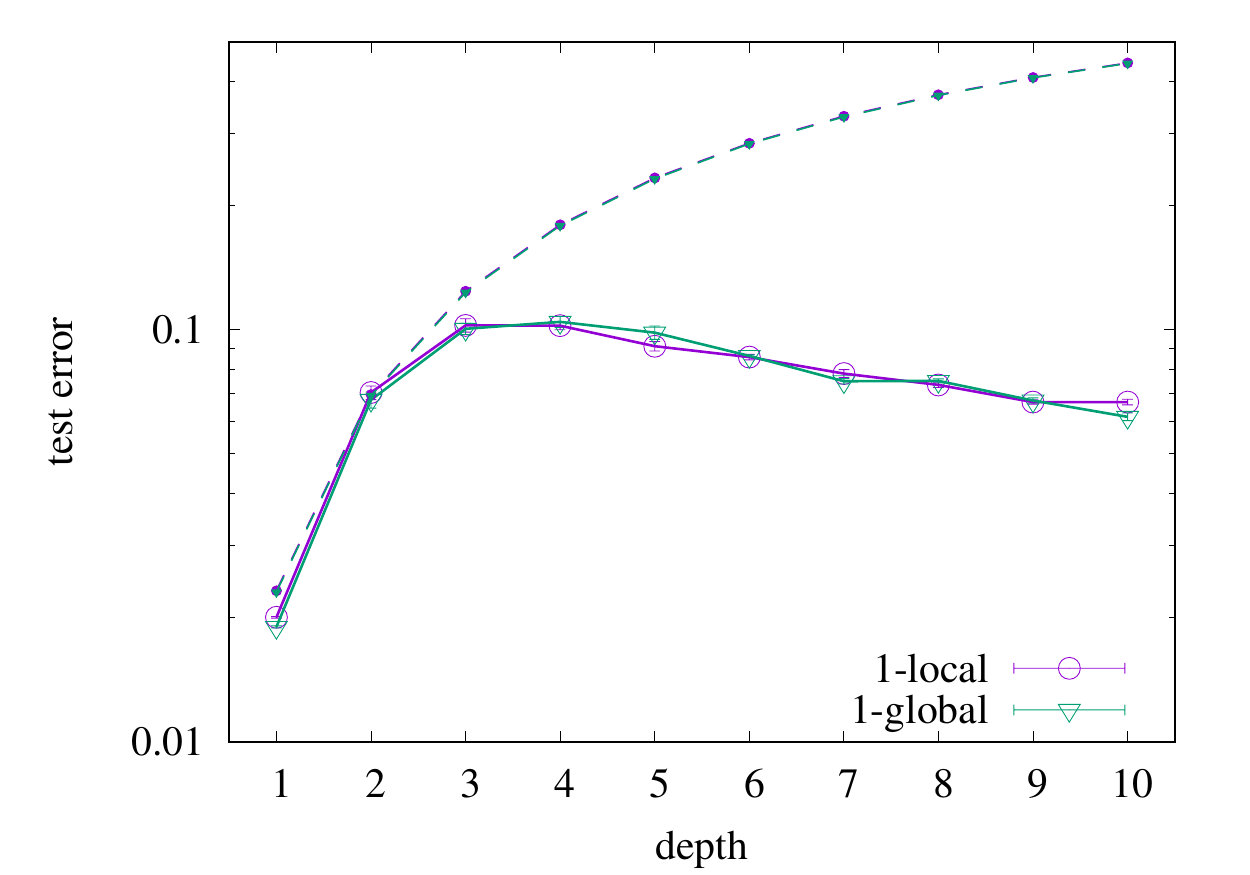}
\caption{Depth dependence of the test error for the 1-local and 1-global target functions.
Solid curves show numerical results in neural networks for various values of the depth with a fixed number of parameters $P=10^8$.
Error bars are typically smaller than symbols.
Dashed lines show numerical results for the NTK. }
\label{fig:dep_1}
\end{figure}

\section{Experimental results}

We now present our experimental results.
First, we show the depth dependence of the test error for the optimized learning rate.
We will see that depth is beneficial for local functions but not for global ones.
This nontrivial interplay between depth and locality is not explained by the NTK.
Next, we investigate the dependence on the learning rate.
We will see that although results for small learning rates agree with those for the NTK, the optimal learning rate is often found in the feature learning regime, which is not described by the NTK.
This result implies the importance of the feature learning in understanding benefits of depth in DNNs.

\begin{figure}[t]
\includegraphics[width=0.45\linewidth]{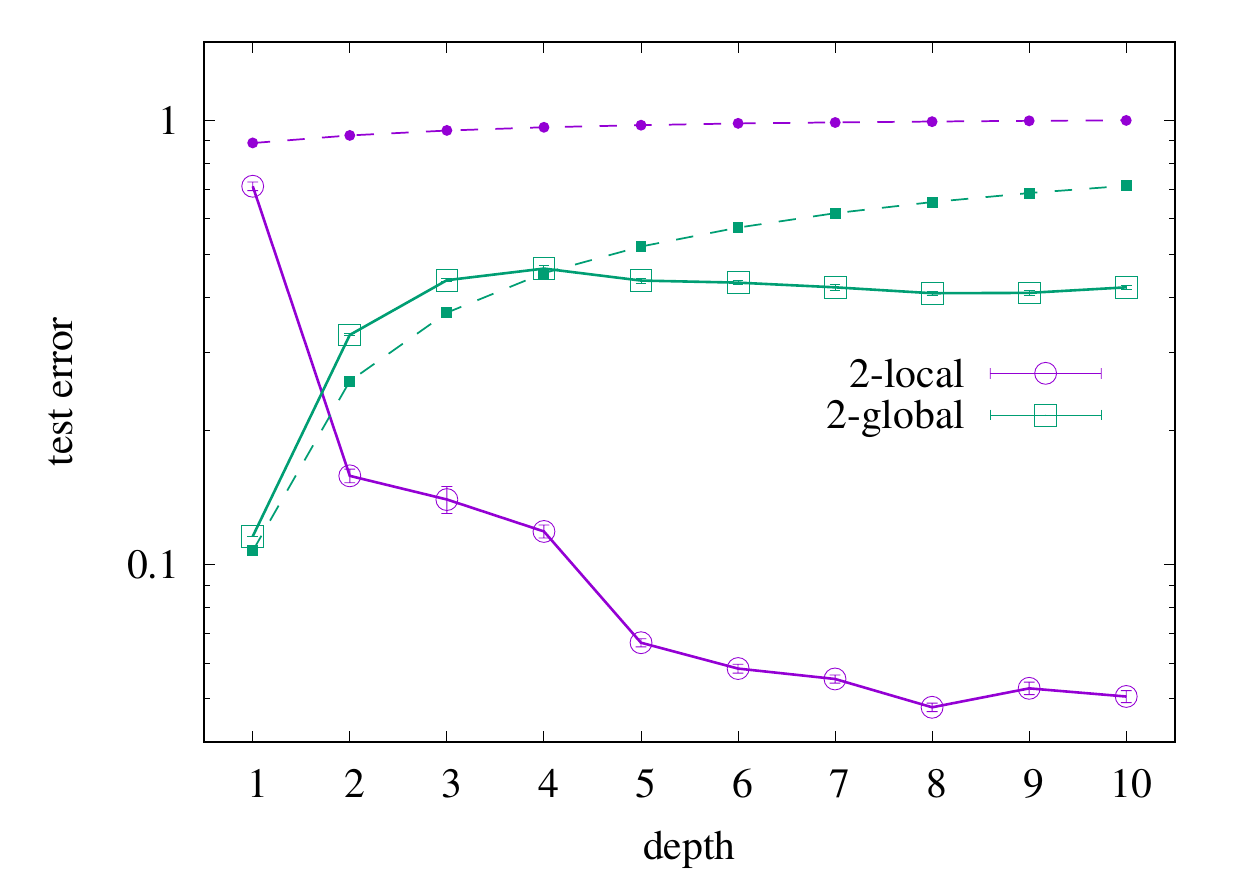}
\includegraphics[width=0.45\linewidth]{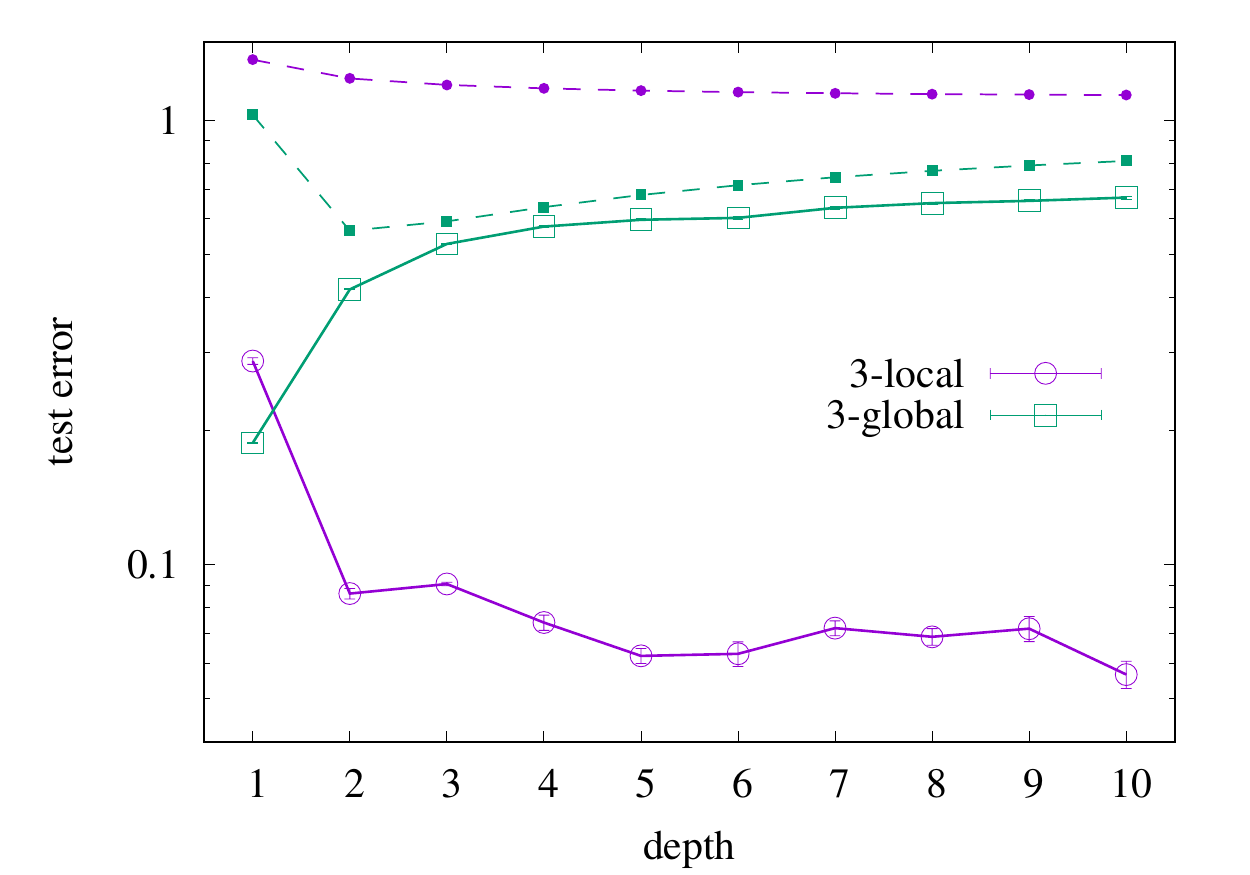}
\caption{Depth dependence of the test error for the $k$-local and $k$-global target functions with $k=2$ (left panel), and $k=3$ (right panel).
Solid curves show numerical results in neural networks for various values of the depth with a fixed number of parameters $P=10^6$.
Error bars are typically smaller than symbols.
Dashed lines show numerical results for the NTK. 
}
\label{fig:dep}
\end{figure}

\subsection{Opposite depth dependence for $k$-local and $k$-global functions}
\label{sec:depth}
We now investigate the depth dependence of the test error.
It turns out that results for linear target functions ($k=1$) qualitatively differ from those for nonlinear ones ($k\geq 2$).
We therefore first show experimental results for $k=1$, and then discuss more intriguing cases of $k\geq 2$.

Numerical results for $k=1$ are shown in Fig.~\ref{fig:dep_1}, where we set $d=1000$, $N=5000$, and $P=10^8$.
We find similar depth dependences for the 1-local and 1-global functions, which indicates that the locality does not matter for linear functions.
We find that a shallow network ($L=1$) outperforms DNNs with $L\geq 2$, although the test error shows non-monotonicity with respect to $L$.
The NTK also predicts that a shallow network is better, but does not reproduce the non-monotonicity.

Results qualitatively change for non-linear target functions with $k\geq 2$.
We show numerical results for $k$-local and $k$-global functions with $k=2$ (left) and 3 (right) in Fig.~\ref{fig:dep}.
The input dimension $d$ and the number  $N$ of training samples are set as $(d,N)=(500,20000)$ for the 2-local function, $(100,10000)$ for the 2-global function, $(100,20000)$ for the 3-local functions, and $(40,20000)$ for the 3-global function.
In all cases, we set $P=10^6$.
Since the values of $d$ and $N$ are chosen differently for different target functions, it is not meaningful to quantitatively compare test errors for different target functions.
Rather, we shall focus on the depth dependence of the test error, which is not sensitive to the choice of $d$ and $N$.

For local functions, the test error for a shallow network of $L=1$ is much higher than that for DNNs.
We find that the test error tends to decrease as the depth $L$ increases, which means that depth is beneficial for learning $k$-local functions.
On the other hand, for global functions, a shallow network shows much better performance than DNNs, which means that depth is rather detrimental to learning global functions.

These results tell us that depth is beneficial even for very simple functions, but does not always help generalization.
Thus, it depends on the locality of target functions (or relevant features within data) whether we should use DNNs.

Remarkably, the NTK is not a good approximation of a neural network at the optimal learning rate (compare solid and dashed lines in Fig.~\ref{fig:dep}), except for the 2-global target function and the 3-global target function with $L\geq 2$.
The interplay between depth and locality is not correctly captured by the NTK.
For example, in the 2-local function, the test error calculated by the NTK increases with depth, although it decreases in neural networks.
In the 3-global function, the NTK seems to be a relatively good approximation for large $L$, but the NTK predicts that a shallow network of $L=1$ generalizes poorer than DNNs, which is not the case in neural networks.

\begin{figure}[t]
\centering
\includegraphics[width=0.45\linewidth]{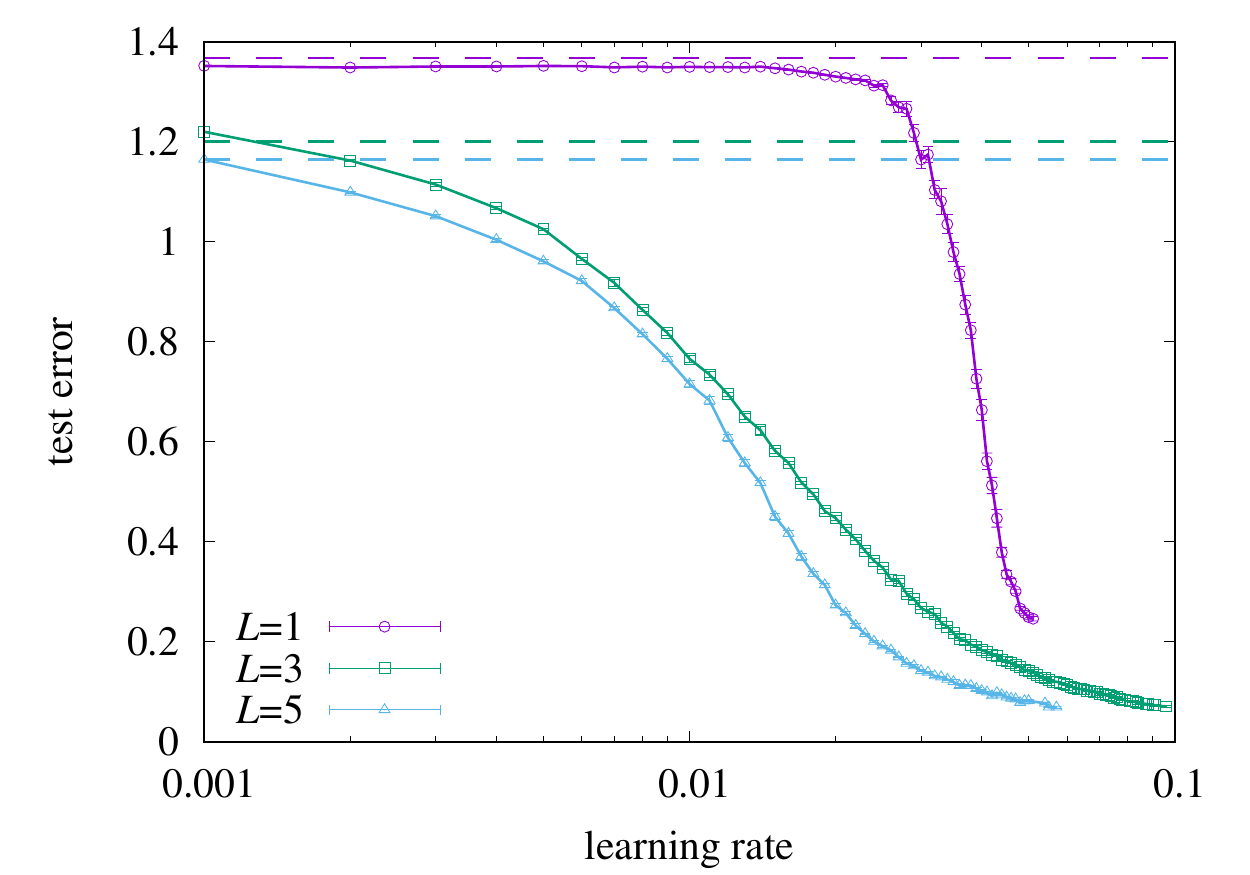}
\includegraphics[width=0.45\linewidth]{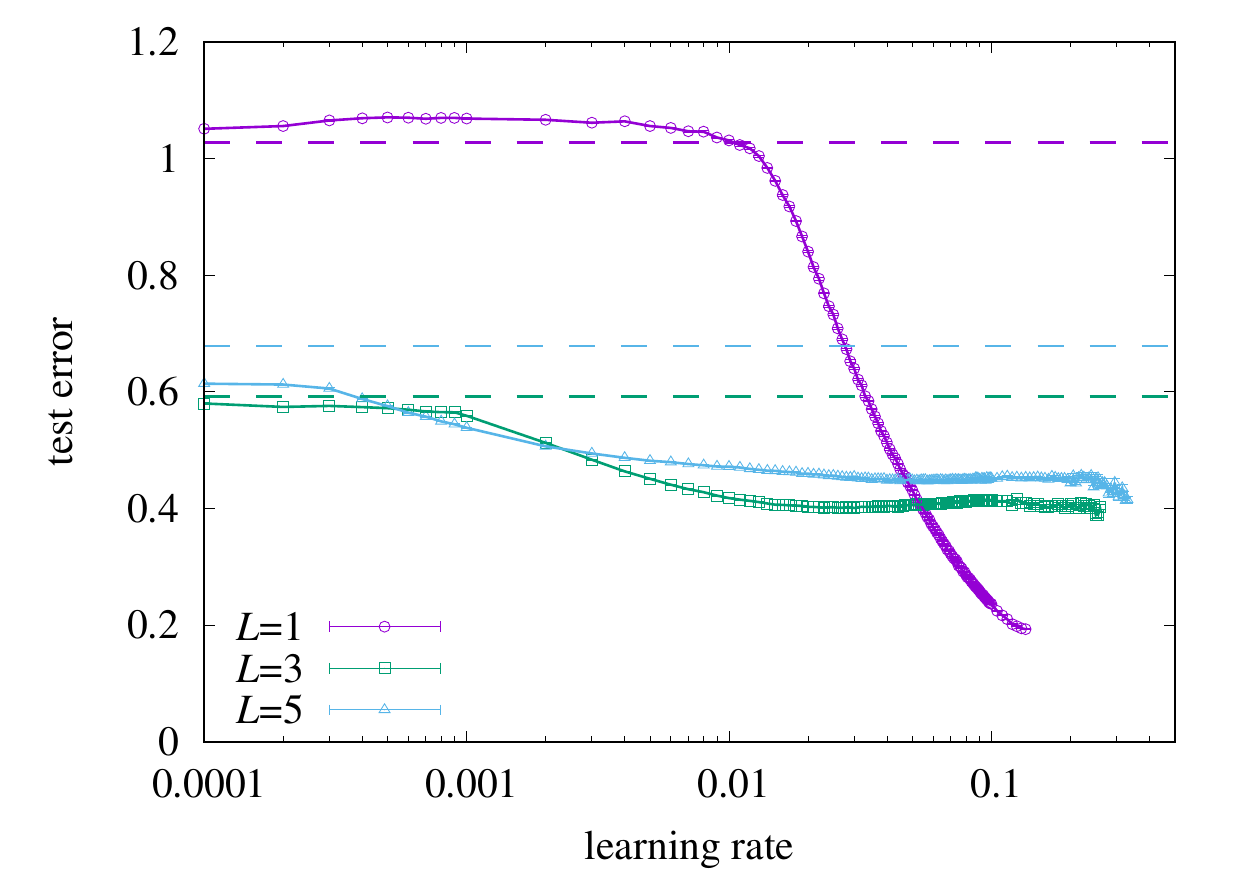}
\caption{Learning-rate dependence of the test error for the 3-local (left panel) and 3-global (right panel) functions.
Solid lines show numerical results in neural networks with the width $h=5000$.
Error bars are typically smaller than symbols.
Dashed lines show test errors calculated by using the NTK. 
}
\label{fig:lr}
\end{figure}

\subsection{Learning rate dependence}
\label{sec:lr}

In \cref{sec:depth}, we find that the NTK does not correctly explain the depth dependence of the test error at an optimal learning rate.
The fact that the NTK describes the lazy learning regime corresponding to small learning rates~\citep{Lee2019} indicates that generalization strongly depends on the learning rate, and the optimal learning rate should be in the feature learning regime~\citep{Lewkowycz2020}.

We shall investigate the learning-rate dependence of the test error.
Numerical results for $k$-local and $k$-global functions with $k=2$ and 3 are shown in Fig.~\ref{fig:lr}.
In Fig.~\ref{fig:lr}, instead of fixing the number $P$ of parameters, we fix the width $h=5000$ of hidden layers.
Each data in Fig.~\ref{fig:lr} is plotted up to the maximum learning rate beyond which the loss value smaller than $10^{-4}$ is not achieved within 2500 epochs (for large learning rates training often fails due to divergence of the network parameters).

We find that the NTK (dashed lines in Fig.~\ref{fig:lr}) is a good approximation in a small learning-rate regime, but not in a large learning-rate regime.
Figure~\ref{fig:lr} also shows that an optimal learning rate is often found in the large learning-rate regime, which is the reason why the NTK cannot capture the interplay between locality and depth.

\subsection{Robustness of the interplay between depth and locality}
\label{sec:robust}
In this section, we show that the interplay between depth and locality observed in \cref{sec:depth} is robust.
We will show that the same depth dependence is observed for (i) noisy labels, (ii) a classification task with the cross-entropy loss, and (iii) more general local and global functions.

\subsubsection{Noisy labels}
\label{sec:noise}

\begin{figure}[t]
\centering
\includegraphics[width=0.45\linewidth]{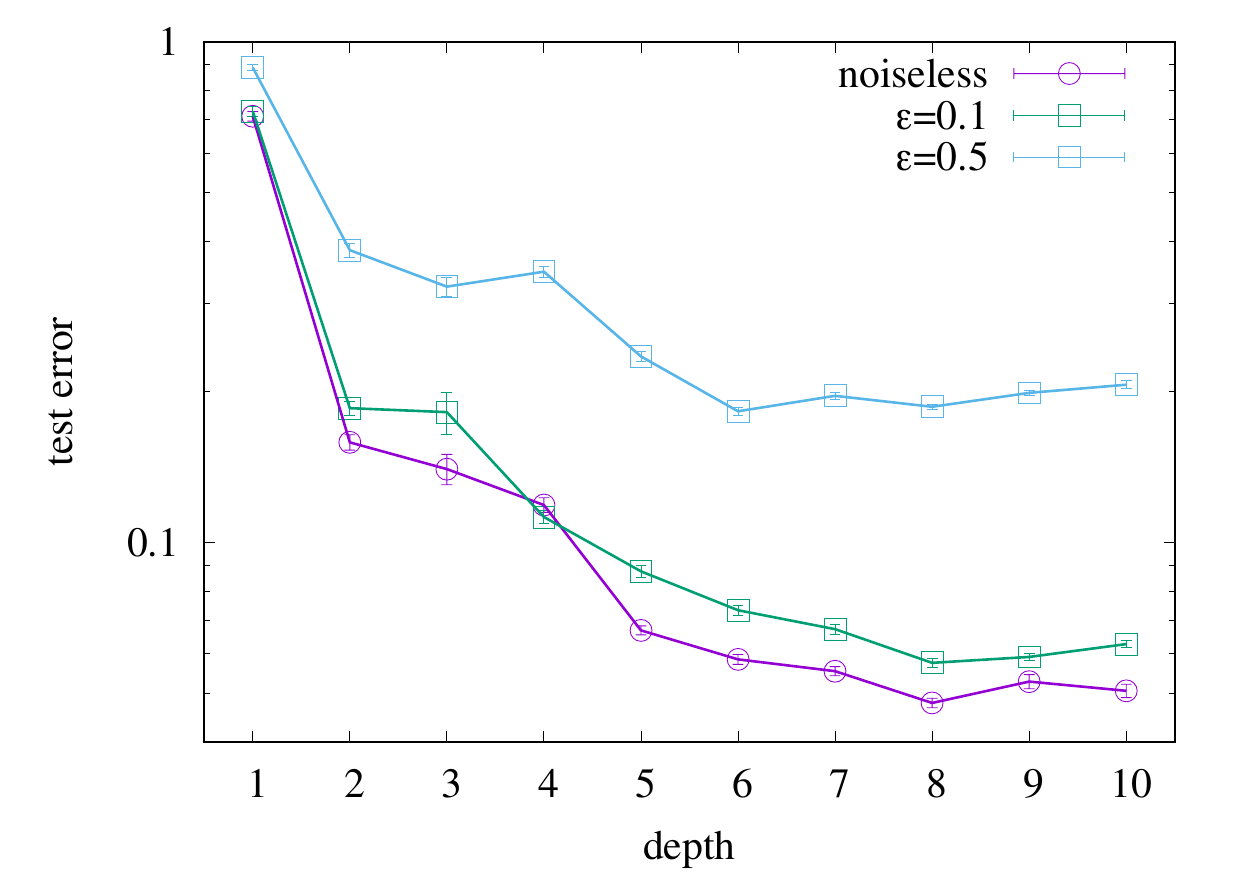}
\includegraphics[width=0.45\linewidth]{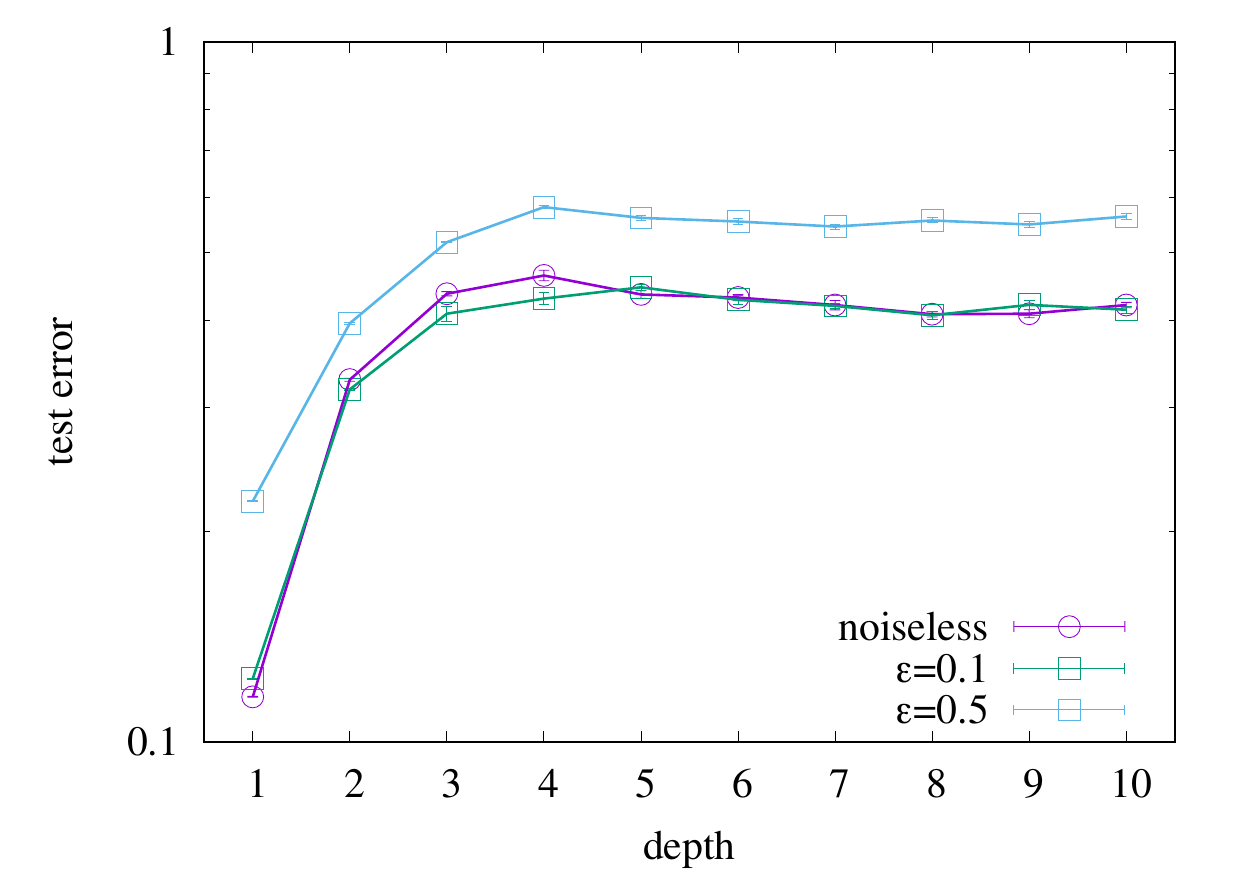}
\caption{Depth dependence of the test error for the $2$-local (left panel) and $2$-global (right panel) functions.
Error bars are typically smaller than symbols.
Numerical results for neural networks with a fixed number $P=10^6$ of the parameters are presented.
}
\label{fig:noise}
\end{figure}

We discuss the effect of noise in the label of the training dataset $\mathcal{D}$: $y^{(\mu)}=f(x^{(\mu)})+\epsilon\xi^{(\mu)}$, where $\xi^{(\mu)}\sim\mathcal{N}(0,1)$ is the Gaussian noise and $\epsilon$ characterizes the noise strength.
The loss function is now given by $L(w)=(1/N)\sum_{\mu=1}^N[\hat{f}(x^{(\mu)},w)-y^{(\mu)}]^2$.
The generalization performance is measured by the test error for noiseless test dataset $\mathcal{D}_\mathrm{test}$, i.e. Eq.~(\ref{eq:error}) is used.
The depth dependences of the test error in the case of the 2-local and the 2-global functions are presented for various values of $\epsilon$ in Fig.~\ref{fig:noise}.
We find that the noise does not change the conclusion that depth is beneficial for local functions but not for global functions.

\subsubsection{Classification}
\label{sec:class}

\begin{figure}[t]
\centering
\includegraphics[width=0.45\linewidth]{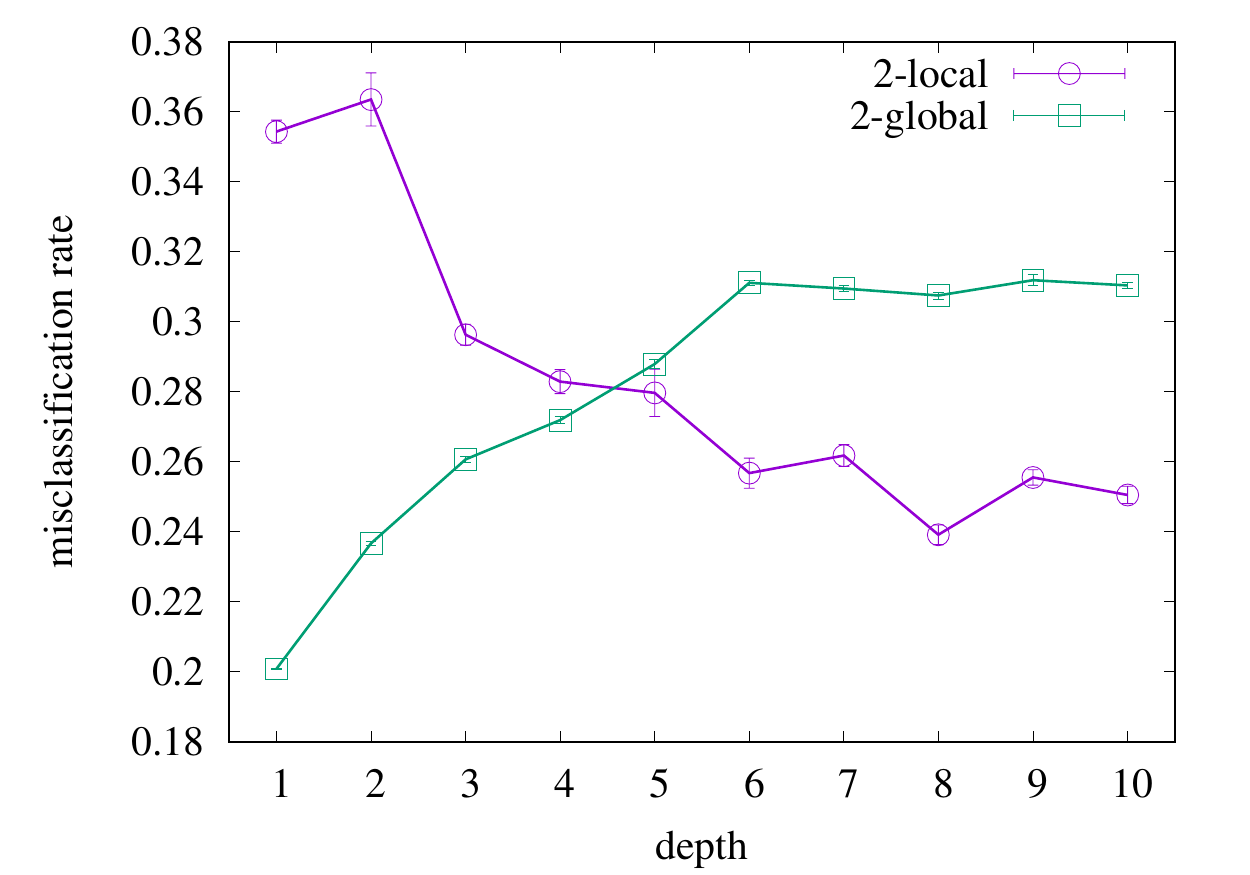}
\includegraphics[width=0.45\linewidth]{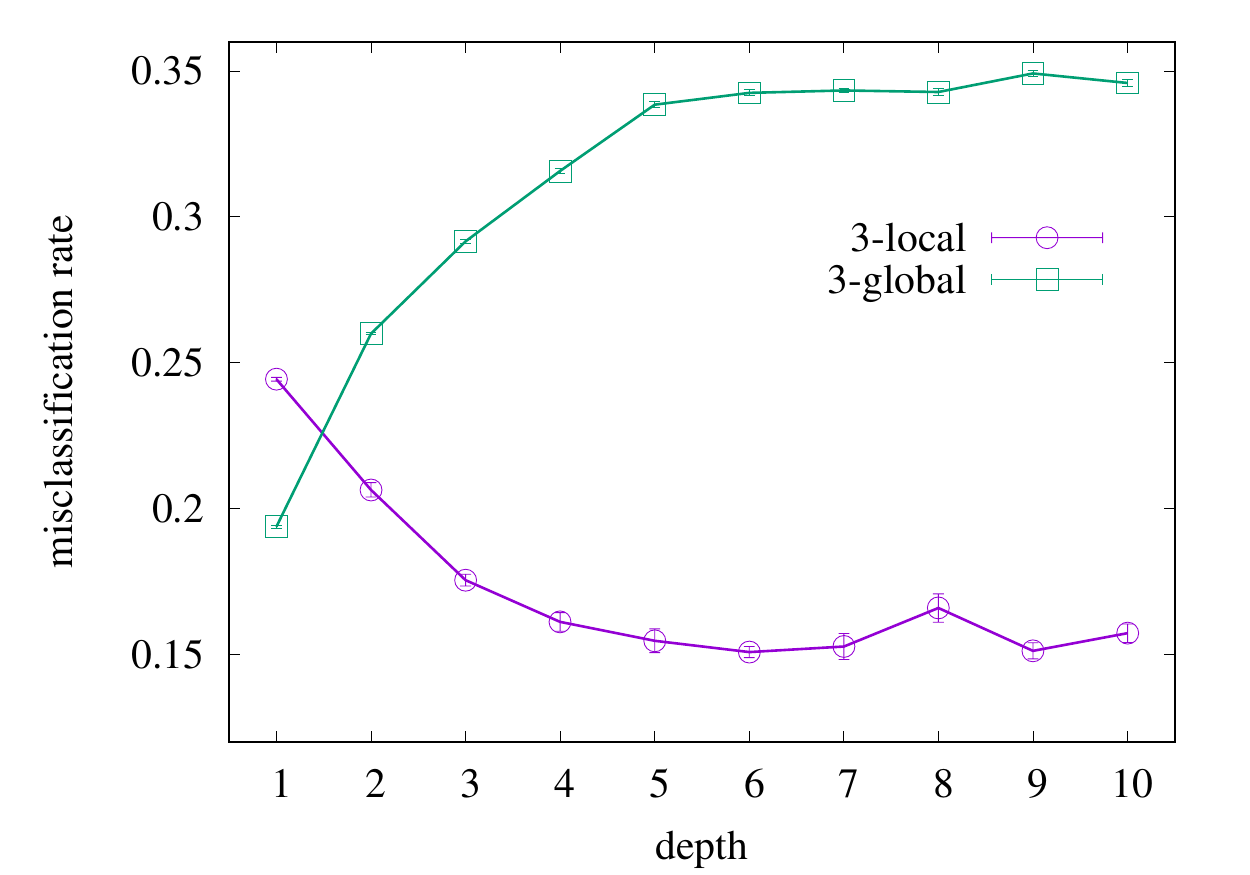}
\caption{Depth dependence of the misclassification rate in a test dataset for a classification task in terms of the sign of the $k$-local and $k$-global functions with $k=2$ (left panel) and $k=3$ (right panel). 
Error bars are typically smaller than symbols.
}
\label{fig:class}
\end{figure}

The opposite depth dependences of the generalization performance for local and global functions are also found in the classification setup.
Now we consider the binary classification problem based on the parity of the $k$-local or $k$-global function $f$.
The label $y$ for an input $x$ is now $y=\mathrm{sgn}[f(x)]$.
We employ the cross-entropy loss as a loss function.
At every 50 epochs, we measure the training accuracy and stop the training if 100\% accuracy is achieved (we have checked that continuing further training does not change the conclusion).
The generalization performance is measured by the misclassification rate for the test dataset.

The depth dependence of the misclassification ratio is shown for $k$-local and $k$-global functions with $k=2$ (left) and $k=3$ (right) in Fig.~\ref{fig:class}.
Here, $(d,N)$ are chosen as (500,10000) for the 2-local function, (100,10000) for the 2-global function, (100,20000) for the 3-local function, and (40,20000) for the 3-global function.
We find that the generalization performance is improved by increasing depth for $k$-local functions, whereas it is worsened for $k$-global functions.

This conclusion is identical to that in the regression setup discussed so far. 
The interplay of depth and locality is not limited to such a specific setup.

\subsubsection{More general local and global functions}
\label{sec:general}

\begin{figure}[t]
\centering
\includegraphics[width=0.45\linewidth]{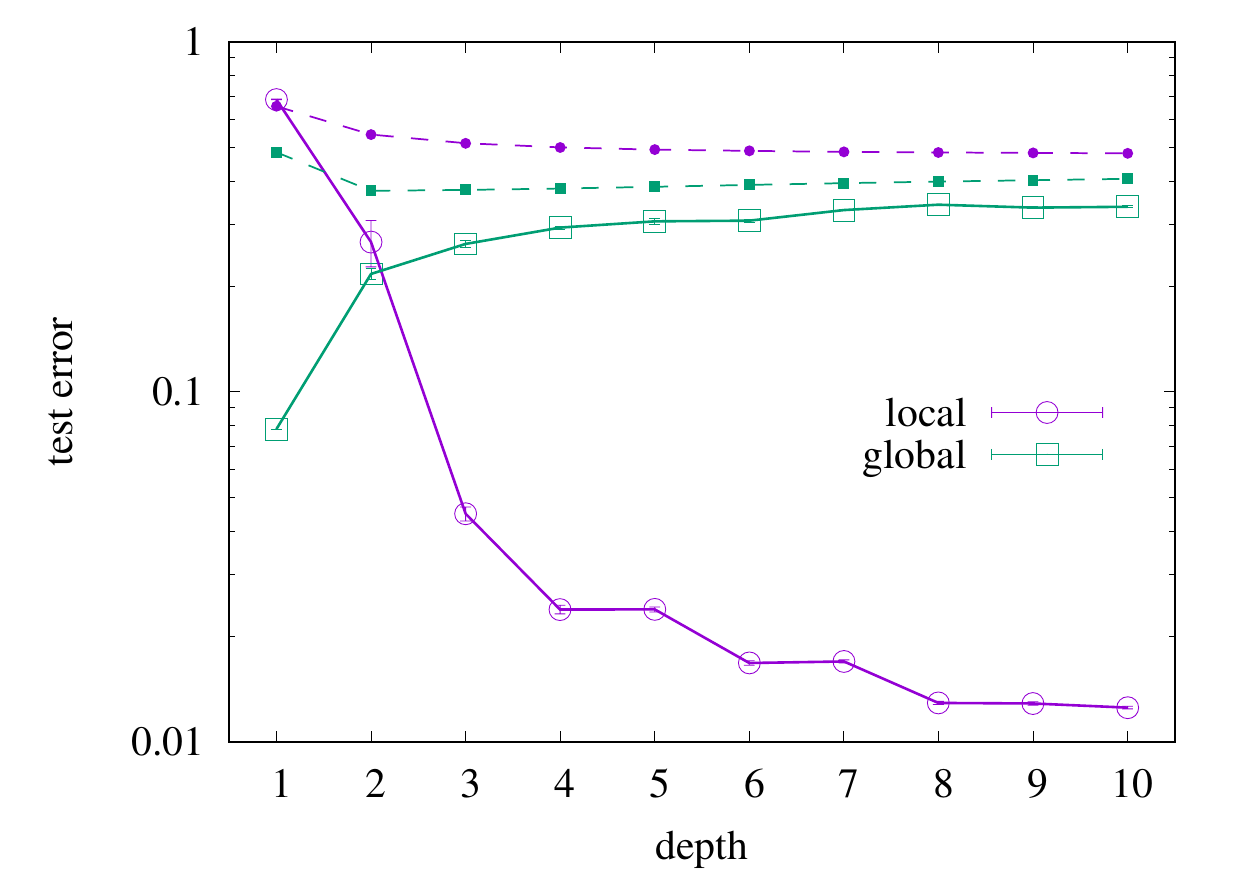}
\includegraphics[width=0.45\linewidth]{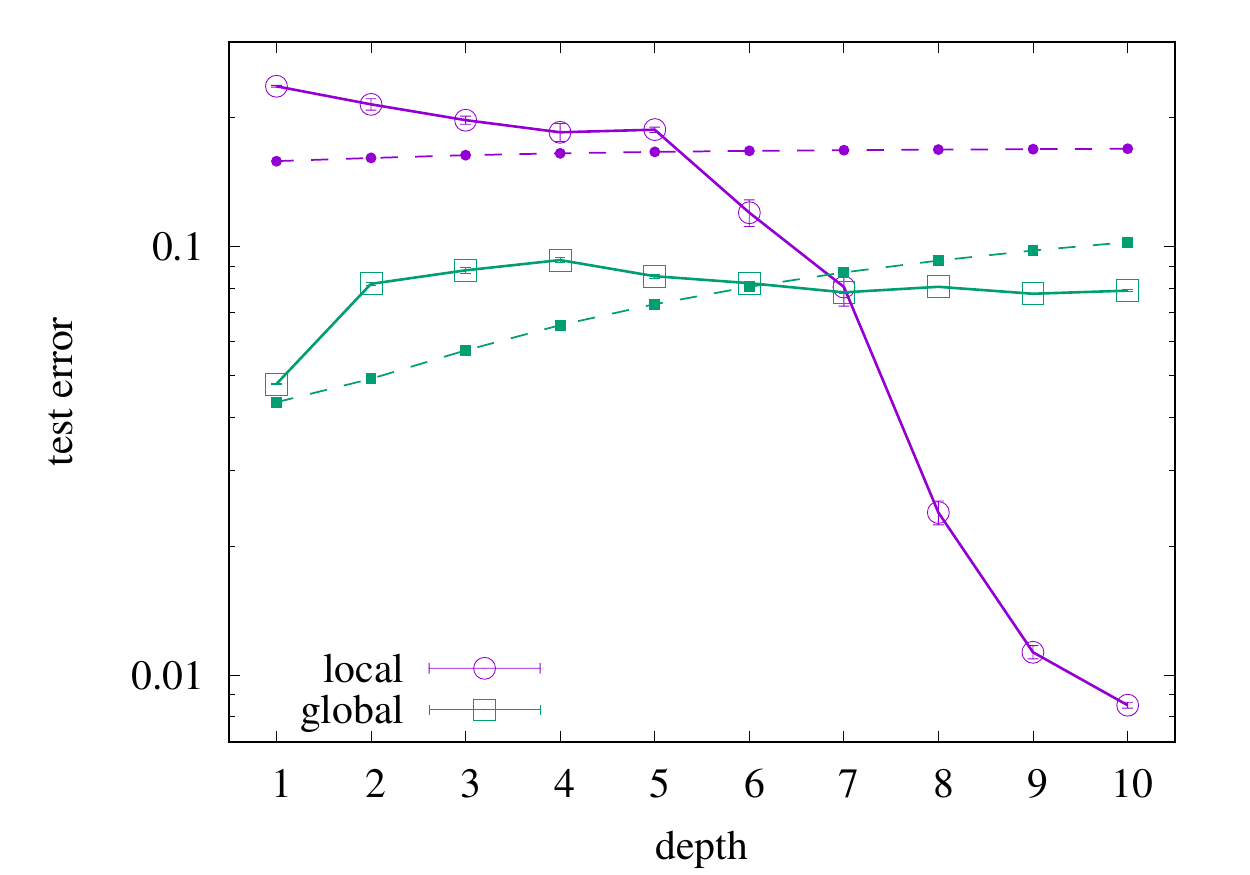}
\caption{Depth dependence of the test error for generalized 2-local and 2-global functions with $g(x_1,x_2)=\sin(2x_1+x_2)$ (left panel) and $g(x_1,x_2)=\tanh(x_1)\sin(x_2)$ (right panel).
Error bars are typically smaller than symbols.
Dashed lines show test errors calculated by the NTK.}
\label{fig:general}
\end{figure}

So far, we have investigated the depth dependence of the generalization performance of DNNs for learning specific $k$-local and $k$-global functions given by Eqs.~(\ref{eq:k-local}) and (\ref{eq:k-global}).
We have seen that depth is beneficial for $k$-local functions, but rather detrimental to learning $k$-global functions.
Here, we extend the notions of $k$-local and $k$-global functions and test whether this conclusion is still true for those extended local and global functions.

Let us introduce a certain (possibly smooth) function $g: \mathbb{R}^k\to\mathbb{R}$ with a positive integer $k$.
We assume $\mathbb{E}_{x_1,x_2,\dots,x_k\sim\mathcal{N}(0,1)}[g(x_1,x_2,\dots,x_k)]=0$.
For a fixed set of $k$ indices $(i_1,i_2,\dots i_k)$ with $1\leq i_1<i_2<\dots<i_k\leq d$, a $k$-local function $f:\mathbb{R}^d\to\mathbb{R}$ is written as
\begin{equation}
f(x)=g(x_{i_1},x_{i_2},\dots,x_{i_k}).
\label{eq:k-local_g}
\end{equation}
The corresponding $k$-global function is defined as
\begin{equation}
f(x)=\frac{1}{\sqrt{d}}\sum_{j=1}^d g(x_{j+i_1},x_{j+i_2},\dots,x_{j+i_k}).
\label{eq:k-global_g}
\end{equation}
Equations~(\ref{eq:k-local}) and (\ref{eq:k-global}) correspond to a simple choice $g(x_1,x_2,\dots,x_k)=x_1x_2\dots x_k$.
Equations~(\ref{eq:k-local_g}) and (\ref{eq:k-global_g}) are thus extensions of Eqs.~(\ref{eq:k-local}) and (\ref{eq:k-global}), respectively.

The interplay between depth and locality observed in \cref{sec:depth} is expected to be true for a more general class of $g$.
It is clearly an important problem to theoretically support this statement.
The fact that the NTK fails to explain this interplay implies that we should investigate generalization performance in the feature learning regime.
It will require new theoretical tools, and hence we postpone it to future studies.
Instead, we shall numerically test whether the same conclusion holds for other two examples of $g$: $g(x_1,x_2)=\sin(2x_1+x_2)$ and $g(x_1,x_2)=\tanh(x_1)\sin(x_2)$.

Numerical results are shown in Fig.~\ref{fig:general} for local and global functions corresponding to these two examples of $g$.
When $g(x_1,x_2)=\sin(2x_1+x_2)$, we set $(d,N)=(200,20000)$ for the local function~(\ref{eq:k-local_g}) and $(d,N)=(30,20000)$ for the global function~(\ref{eq:k-global_g}).
When $g(x_1,x_2)=\tanh(x_1)\sin(x_2)$, we set $(d,N)=(500,20000)$ for the local function~(\ref{eq:k-local_g}) and $(d,N)=(100,20000)$ for the global function~(\ref{eq:k-global_g}).
We find that, in both cases, DNNs outperform a shallow network when the target function is local, whereas a shallow network outperforms DNNs when the target function is global.
In this way, the interplay between depth and locality that is observed in \cref{sec:depth} is robust against the change of the function $g$.

\section{Conclusion}
We have seen that depth is beneficial for local functions but not for global functions in an overparameterized regime.
In previous works~\citep{Telgarsky2016, Poole2016, Bianchini2014, Montufar2014}, benefits of DNNs have been partially attributed to their high expressivity, which indicates that benefits of depth are expected to be evident for highly complex target functions.
However, our $k$-local functions given by Eq.~(\ref{eq:k-local}) are very simple, which clearly shows that benefits of depth presented in our work are not due to high expressive powers of DNNs.

It would also be an interesting observation that depth is rather detrimental to learning global target functions.
While there are many studies on benefits of depth, it is also important to figure out when depth is disadvantageous.

As is demonstrated in \cref{sec:robust}, the above conclusion is robust against some changes of setting.
It indicates that some underlying fundamental mechanism exists.
In particular, results in \cref{sec:general} show that the interplay of depth and locality is not a special property of specific functions of Eqs.~(\ref{eq:k-local}) and (\ref{eq:k-global}).
Rather, this interplay will be a general property in a certain class of local and global functions written in the form of Eqs.~(\ref{eq:k-local_g}) and (\ref{eq:k-global_g}), respectively.
It is an open problem to theoretically understand such a fundamental mechanism.

Since this interplay is not observed in the lazy learning regime, in which the NTK is an adequate theoretical tool, we should theoretically investigate the feature learning regime to understand the mechanism behind it.
A new theoretical tool will be required, and so we leave it as an important future problem.

Here, we have to be content with just presenting an intuitive argument towards this direction.
Since information on an input vector is lost by propagating through the network layer by layer~\citep{Poole2016, Schoenholz2017}, it is expected that DNNs are suited for local target functions, in which most elements of an input vector are irrelevant (we should be willing to throw away information on the data).
By utilizing the chaoticity of information processing in DNNs~\citep{Poole2016, Schoenholz2017}, we can successively amplify a local change of an input vector through hidden layers while throwing away irrelevant information.
In contrast, global target functions depend on all elements of an input vector, and hence information on the input should be kept at the output layer.
In that case, depth can rather be detrimental to generalization.

The above argument is still primitive.
It is a challenging theoretical problem to establish a precise mathematical theory.

\bibliography{apsrevcontrol,deep_learning,physics}
\bibliographystyle{apsrev4-2}
\clearpage
\appendix
\section{Explicit expression of the NTK}
\label{sec:formula_NTK}

We consider a network whose biases $\{b^{(\ell)}\}$ and weights $\{w^{(\ell)}\}$ are randomly initialized as $b_i^{(\ell)}=\beta \tilde{b}_i^{(\ell)}$ with $\tilde{b}_i^{(\ell)}\sim\mathcal{N}(0,1)$ and $w_{ij}^{(\ell)}=\sqrt{2/n_{\ell-1}}\tilde{w}_{ij}^{(\ell)}$ with $\tilde{w}_{ij}^{(\ell)}\sim\mathcal{N}(0,1)$ for every $\ell$, where $n_\ell$ is the number of neurons in the $\ell$th layer, i.e., $n_0=d$, $n_1=n_2=\dots=n_L=h$.
In the infinite-width limit $h\to\infty$, the pre-activation $f^{(\ell)}=w^{(\ell)}z^{(\ell-1)}+b^{(\ell)}$ at every hidden layer tends to an i.i.d. Gaussian process with covariance $\Sigma^{(\ell-1)}:\mathbb{R}^d\times\mathbb{R}^d\to\mathbb{R}$ which is defined recursively as
\begin{equation}
\left\{
\begin{split}
&\Sigma^{(0)}(x,x')=\frac{x^\mathrm{T}x'}{d}+\beta^2; \\
&\Lambda^{(\ell)}(x,x')=\mqty(\Sigma^{(\ell-1)}(x,x) & \Sigma^{(\ell-1)}(x,x') \\ \Sigma^{(\ell-1)}(x',x) & \Sigma^{(\ell-1)}(x',x')); \\
&\Sigma^{(\ell)}(x,x')=2\mathbb{E}_{(u,v)\sim\mathcal{N}(0,\Lambda^{(\ell)})}\qty[\varphi(u)\varphi(v)]+\beta^2
\end{split}
\right.
\end{equation}
for $\ell=1,2,\dots, L$.
We also define 
\begin{equation}
\dot{\Sigma}^{(\ell)}(x,x')=2\mathbb{E}_{(u,v)\sim\mathcal{N}(0,\Lambda^{(\ell)})}\qty[\dot{\varphi}(u)\dot{\varphi}(v)],
\end{equation}
where $\dot{\varphi}$ is the derivative of $\varphi$.
The NTK is then expressed as
\begin{equation}
\Theta^{(L)}(x,x')=\sum_{\ell=1}^{L+1}\qty(\Sigma^{(\ell-1)}(x,x')\prod_{\ell'=\ell}^{L+1}\dot{\Sigma}^{(\ell')}(x,x')).
\label{eq:NTK}
\end{equation}
The derivation of this formula is given by \citet{Arora2019}.

Using the ReLU activation function $\varphi(u)=\max\{u,0\}$, we can further calculate $\Sigma^{(\ell)}(x,x')$ and $\dot{\Sigma}^{(\ell)}(x,x')$~\citep{Lee2019}, obtaining
\begin{equation}
\Sigma^{(\ell)}(x,x')=\frac{\sqrt{\det\Lambda^{(\ell)}}}{\pi}+\frac{\Sigma^{(\ell-1)}(x,x')}{\pi}\left[\frac{\pi}{2}+\arctan\left(\frac{\Sigma^{(\ell-1)}(x,x')}{\sqrt{\det\Lambda^{(\ell)}}}\right)\right]+\beta^2
\label{eq:Sigma}
\end{equation}
and
\begin{equation}
\dot{\Sigma}^{(\ell)}(x,x')=\frac{1}{2}\left[1+\frac{2}{\pi}\arctan\left(\frac{\Sigma^{(\ell-1)}(x,x')}{\sqrt{\det\Lambda^{(\ell)}}}\right)\right].
\label{eq:dot_Sigma}
\end{equation}
For $x=x'$, we obtain $\Sigma^{(\ell)}(x,x)=\Sigma^{(0)}(x,x)+\ell\beta^2=\|x\|^2/d+(\ell+1)\beta^2$.
By solving Eqs.~(\ref{eq:Sigma}) and (\ref{eq:dot_Sigma}) iteratively, we obtain the NTK in Eq.~(\ref{eq:NTK}).\footnote{When $\beta=0$ (no bias), the equations are further simplified as $\Sigma^{(\ell)}=\frac{\|x\|\|x'\|}{d}\cos\theta^{(\ell)}$ and $\dot{\Sigma}^{(\ell)}=1-\frac{\theta^{(\ell-1)}}{\pi}$, where $\theta^{(0)}\in[0,\pi]$ is the angle between $x$ and $x'$, and $\theta^{(\ell)}$ is determined recursively by $\cos\theta^{(\ell)}=\frac{1}{\pi}\left[\sin\theta^{(\ell-1)}+(\pi-\theta^{(\ell-1)})\cos\theta^{(\ell-1)}\right]$.}

\end{document}